%% file: main.tex
\documentclass[final]{cvpr}

\usepackage{times}
\usepackage{epsfig}
\usepackage{graphicx}
\usepackage{amsmath}
\usepackage{amssymb}
\usepackage[utf8]{inputenc}
\usepackage{mathtools}
\usepackage{xcolor}

\usepackage{multirow}
\usepackage{booktabs}
\usepackage{cuted}
\usepackage{algorithm}
\usepackage{algorithmic}
\usepackage[font=small,labelfont=bf,tableposition=top]{caption}

\usepackage[pagebackref=true,breaklinks=true,letterpaper=true,colorlinks,bookmarks=false]{hyperref}

\def\cvprPaperID{7574} 
\def\confYear{CVPR 2021}
\begin{document}
\title{PiCIE: Unsupervised Semantic Segmentation using Invariance and Equivariance in Clustering}
\author{Jang Hyun Cho$^1$
\quad\quad\quad
Utkarsh Mall$^2$
\quad\quad\quad
Kavita Bala$^2$
\quad\quad\quad
Bharath Hariharan$^2$\\
$^1$University of Texas at Austin\quad\quad\quad\quad $^2$Cornell University\\
}
\maketitle

\setcounter{footnote}{0}

\begin{strip}
\centering\noindent
\includegraphics[width=0.85\linewidth, trim={0 0 0 2cm}]{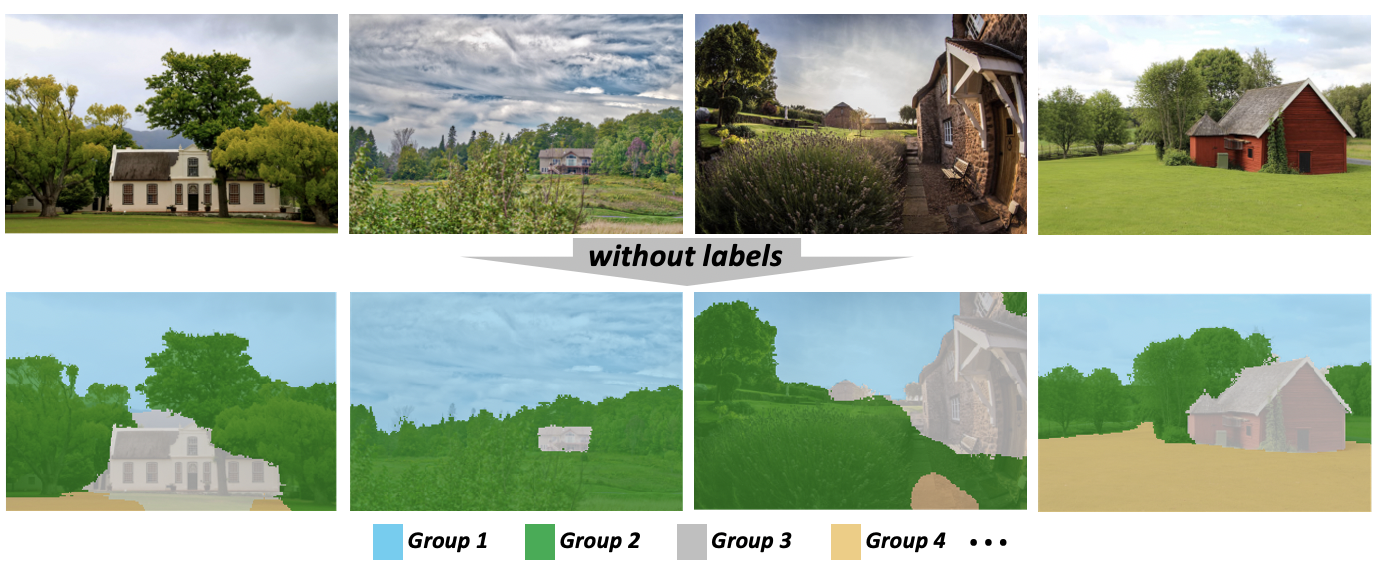}
\captionof{figure}{ From these unannotated images, we would like a recognition system to discover the concepts of \emph{house, grass, trees} and \emph{sky}, and segment each image accordingly without any supervision. }
\label{fig:intro}
\vspace{-3mm}
\end{strip}

\begin{abstract}
\vspace{-4mm}
We present a new framework for semantic segmentation without annotations via clustering. Off-the-shelf clustering methods are limited to curated, single-label, and object-centric images yet real-world data are dominantly uncurated, multi-label, and scene-centric. We extend clustering from images to pixels and assign separate cluster membership to different instances within each image. However, solely relying on pixel-wise feature similarity fails to learn high-level semantic concepts and overfits to low-level visual cues. We propose a method to incorporate geometric consistency as an inductive bias to learn invariance and equivariance for photometric and geometric variations. With our novel learning objective, our framework can learn high-level semantic concepts. Our method, \textbf{PiCIE} (\textbf{Pi}xel-level feature \textbf{C}lustering using \textbf{I}nvariance and \textbf{E}quivariance), is the first method capable of segmenting both \emph{things} and \emph{stuff} categories without any hyperparameter tuning or task-specific pre-processing. Our method largely outperforms existing baselines on COCO~\cite{coco} and Cityscapes~\cite{cityscapes} with \textbf{+17.5} Acc. and \textbf{+4.5} mIoU. We  show that PiCIE gives a better initialization for standard supervised training. The code is available at \url{https://github.com/janghyuncho/PiCIE}.
\end{abstract}


\begin{figure*}
\centering\noindent
\includegraphics[width=\linewidth]{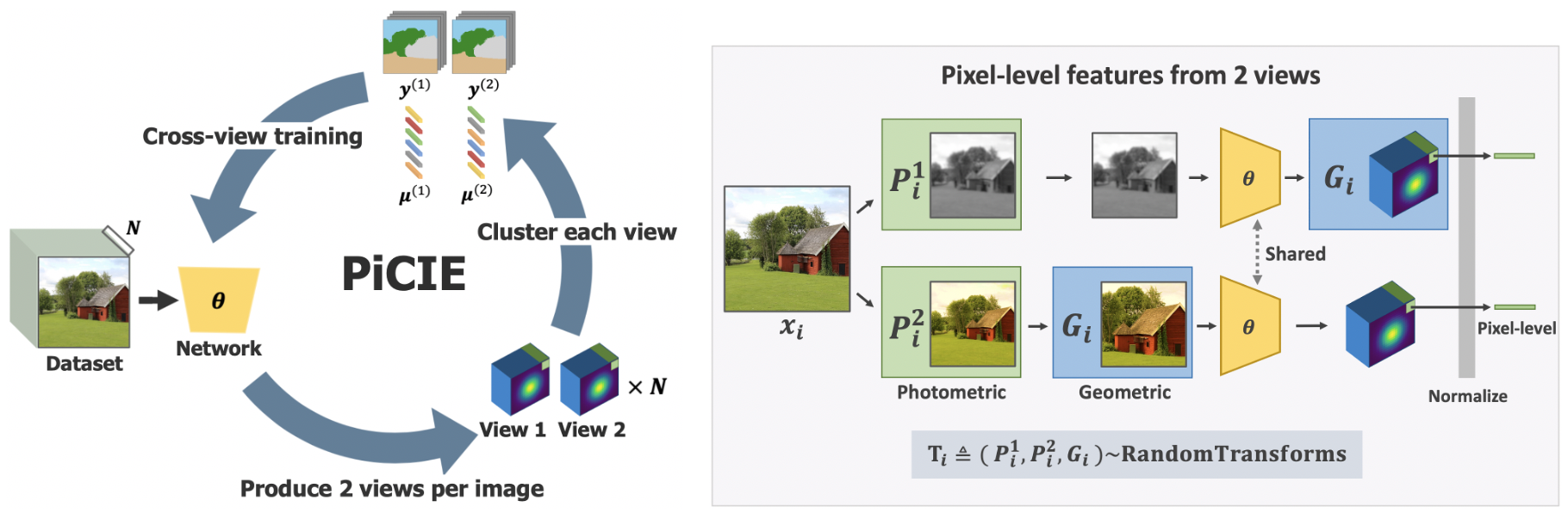}
\vspace{-6mm}
\caption{{PiCIE overview (\textbf{left}) and illustration of multi-view feature computation (\textbf{right}). More details in Sec.~\ref{inv_eqv}. } 
}
\vspace{-4mm}
\label{fig:pipeline}
\end{figure*}

\section{Introduction}
Unsupervised learning from a set of unlabelled images has gained large popularity, but still is mostly limited to single-class, object-centric images. 
Consider the images shown in Figure~\ref{fig:intro} (top). 
Given a collection of these and other unlabeled images, can a machine discover the concepts of ``grass'', ``sky'', ``house'' and ``trees'' from \emph{each} image?
Going further, can it identify \emph{where} in each image each concept appears, and \emph{segment} it out?

A system that is capable of such \emph{unsupervised semantic segmentation} can then automatically discover classes of objects with their precise boundaries, thus removing the substantial cost of collecting and labeling datasets such as COCO. It might even discover objects, materials and textures that an annotator may not know of \emph{a priori}. This can be particularly useful for analyzing novel domains: for example, discovering new kinds of visual structures in satellite imagery. The ability of the system to discover and segment out unknown objects may also prove useful for robots trying to manipulate these objects in the wild.

However, while unsupervised semantic segmentation might be useful, it is also challenging.
This is because it combines the problem of class discovery with the challenge of exhaustive pixel labeling.
Recent progress in self-supervised and unsupervised learning suggests that recognition systems can certainly discover \emph{image-level} classes.
However, image-level labeling is easier since the network can simply rely on just a few distinctive, stable features and discard the rest of the image.
For example, a recognition system might be able to group all four images of Figure~\ref{fig:intro} together simply by detecting the presence of roof tiles in each image, and ignoring everything else in the images.
In contrast, when \emph{segmenting} the image, no pixel can be ignored; whether it is a distinct object (\emph{thing}) or a background entity (\emph{stuff}), \emph{each and every pixel must be recognized and accurately characterized} in spite of potentially large intra-class variation.
As such, very little prior work has tried to tackle this problem of discovering semantic segmentations, with results limited to extremely coarse \emph{stuff} segmentation. 

In this paper, we take a step towards a practically useful unsupervised semantic segmentation system: we present an approach that is able to segment out all pixels, be they \emph{things} or \emph{stuff}, at a much finer granularity than prior art. 
Our approach is based on a straightforward objective that codifies only two common-sense constraints. 
First, pixels that have a similar appearance (i.e., they cluster together in a learned feature space) should be labeled similarly and vice versa.
Second, pixel labels should be \emph{invariant} to color space transformations and \emph{equivariant} to geometric transformations.
Our results show that using these two objectives alone, we can train a ConvNet based semantic segmentation system \emph{end-to-end} without any labels.

We find that in spite of its simplicity, our approach far outperforms prior work on this task, \emph{more than doubling the accuracy} of prior art (Figure~\ref{fig:intro}, bottom).
Our clustering-based loss function (the first objective above) leads to a much simpler and easier learning problem compared to prior work, which instead tries to learn parametric pixel classifiers.
But the invariance and equivariance objectives are key.
They allow the convolutional network to connect together pixels across scale, pose and color variation, something that prior systems are unable to do.
This increased robustness to invariance also allows our approach to effectively segment \emph{objects}.
We vindicate these intuitions through an ablation study, where we find that each of these contributes significant improvements in performance.

In sum, our results show that convolutional networks can learn to not only discover image-level concepts, but also semantically parse images without any supervision. This opens the door to true large-scale discovery, where such a trained network can automatically \emph{surface} new classes of objects, materials or textures from only an unlabeled, uncurated dataset.

\section{Related Work}

\vspace{-2mm}
\paragraph{Learning for clustering.} Using deep neural networks to learn cluster-friendly embedding space has been widely studied~\cite{dc1, dc2, dc3, odc, dec, dec_improved, friend1}. DEC~\cite{dec} 
and IDEC~\cite{dec_improved} train embedding function by training autoencoder (AE)~\cite{sae} with reconstruction loss.
DeepCluster and its variants~\cite{dc1, dc2, odc} explicitly cluster the feature vectors of the entire dataset using k-means~\cite{km} in order to assign \emph{pseudo-labels} to each data point, and then train an encoder network.
All these methods share a common philosophy that iterative optimization of clustering loss improves the feature space to account for high-level visual similarity. 

Apart from a representation learning perspective, there have been a number of recent works that tackle classification without labels by clustering data points~\cite{dec, dtc,sdtc, iic, scan, selflabel}. IIC~\cite{iic}, SeLa~\cite{selflabel} and other works~\cite{scan, mi1, mi2, pseudosemi} maximize mutual information between two versions of soft cluster assignments from a single image. Maximizing mutual information prevents the network from falling into a degenerate solution, but effectively enforces uniform distribution over clusters. Hence, unsupervised clustering is expected to work only with well-balanced datasets such as MNIST~\cite{mnist} and CIFAR~\cite{cifar}. Recent works~\cite{scan, selflabel} tested on larger-scale datasets such as ImageNet~\cite{imagenet}, still assume a balanced set of single-class, object-centric images. Since these methods do not explicitly perform clustering on data, they are called \emph{implicit clustering} methods, contrary to \emph{explicit clustering}~\cite{dec, dec_improved, dc1, dc2, odc, dc4, dc5, dc6, dc8}. 

\vspace{-2mm}
\paragraph{Segmentation without labels. } In clustering, each data point is assumed to be semantically homogeneous. This condition is invalid when images contain more than one semantic class, such as scene-centric datasets~\cite{pascal, coco, cityscapes, lvis}. In fact, the majority of \emph{common} images are not object-centric, and therefore one cannot simply use off-the-shelf clustering methods to obtain semantic understanding of an arbitrary dataset. The problem reduces to \emph{semantic segmentation} by clustering pixel-level features.

There has been a number of recent attempts to semantic segmentation without labels. IIC~\cite{iic} simply extends mutual information-based clustering to pixel-level representation by outputting a probability map over image pixels. AC~\cite{ac} uses an  autoregressive model~\cite{pixelcnn} to obtain probabilities of pixels over categories, which then maximizes mutual information across two different ``orderings'' of autoregression. Both works are limited to \emph{stuff} categories due to the  following two reasons. First, a mixture of \emph{stuff} and \emph{things} categories introduces severe data imbalance since there are far more \emph{stuff} pixels than \emph{things} pixels in real-world images. Such imbalance leads the mutual information maximization to forcibly balance the size of clusters and hence leads to noisy representation as major classes (\emph{stuff} categories) subsume minor classes (\emph{things} categories). Second, each method exploits the \emph{local spatial consistency} condition; a pixel needs to be semantically (and visually) consistent with its neighboring pixels. This condition is only valid with \emph{stuff} categories (e.g., \emph{sky}) and not often true with \emph{things} categories. Other methods~\cite{redrawing,ganss} based on GANs~\cite{gan,pgan} learn to generate foreground masks of a given image, but are limited to a single-category setting. Our method is free from such assumptions and the results show that our method is capable of segmenting both \emph{stuff} and \emph{things} categories together well with uncurated images.

\vspace{-2mm}
\paragraph{Equivariance learning.} Equivariance learning has been studied in object and keypoints tracking~\cite{eqv4, track1, tracking2, tracking3}, facial landmark detection~\cite{face1,face2, face3}, and keypoint detection~\cite{eqv1, eqv2, eqv3, keypoint1, reviewer_asked}. The central idea in these works is to train a model that predicts consistent key points between two images,
 with the underlying assumption that two images share a common instance. This enables unsupervised learning of semantically consistent and geometrically structured representation learning. The general objective is to directly minimize the L2 distance between two feature vectors that correspond to the semantically equivalent locations on images. However, using MSE loss with clustering is often sensitive to the choice of hyper-parameters, which is often infeasible or prone to overfit in unsupervised setting. Furthermore, individual feature vector may contain noisy low-level visual cues which can overwhelm the gradient flow during back-propagation. Our method instead learns equivariance by enforcing consistent clustering assignments between two views and hence only cluster-centered visual cues affect the loss (detail in Sec.~\ref{inv_eqv}).

\section{PiCIE}

We are given an \emph{uncurated, unlabeled} dataset of images taken from some domain $\mathcal{D}$.
On this dataset, we want to discover a set of visual classes $\mathcal{C}$ and learn a semantic segmentation function $f_\theta$. When provided an unseen image from $\mathcal{D}$, $f_\theta$ should be able to assign every pixel a label from the set of classes $\mathcal{C}$.

We formulate this task of unsupervised image segmentation as pixel-level clustering, where every pixel is assigned to a cluster.
Clustering typically requires a good feature space, but no such feature representation exists \emph{a priori}.
We therefore propose an approach that learns the feature representation jointly with the clustering.
The overall pipeline of \textbf{PiCIE}, which stands for \textbf{Pi}xel-level feature \textbf{C}lustering using \textbf{I}nvariance and \textbf{E}quivariance, is depicted in Figure~\ref{fig:pipeline}. 
We describe our approach below.

\subsection{A baseline clustering approach}
\label{overview}
We begin with prior work that learns a neural network end-to-end for clustering unlabeled images into image-level classes~\cite{dc1, dc2, dec, dec_improved, jule}.
The key issue tackled in these papers is that clustering images into classes requires strong feature representations, but for training strong feature representations one needs class labels.
To solve this chicken-and-egg problem, the simplest solution is the one identified by DeepCluster~\cite{dc1}: alternate between clustering using the current feature representation, and use the cluster labels as pseudo-labels to train the feature representation.
One can follow a similar strategy for the unsupervised semantic segmentation task.
The only difference is that we need to use an embedding function $f_\theta$ that produces a feature map, producing a feature vector for every pixel.
The classifier must also operate on individual pixels.
One can then alternate between clustering the pixel feature vectors to get pixel pseudo-labels, and using these pseudo-labels to train the pixel feature representation.

Concretely, suppose we have a set of unlabeled images $x_i, i=1, \ldots, n$.
Suppose our embedding, denoted by $f_\theta$ produces a feature tensor $f_\theta(x)$.
This yields a feature representation for every pixel $p$ in the image $x$.
Denote by $f_\theta(x)[p]$ this pixel-level feature representation.
Denote by $g_{\mathbf{w}}(\cdot)$ a classifier operating on these pixel feature vectors.
Then our baseline approach alternates between two steps:
\begin{enumerate}
\item Use the current embedding and k-means to cluster the pixels in the dataset.
\begin{align}
\min_{\mathbf{y}, \boldsymbol{\mu}} \sum_{i,p} \| f_\theta(x_i)[p] - \mu_{y_{ip}}\|^2
\end{align}
where $y_{ip}$ denotes the cluster labels of the $p$-th pixel in the $i$-th image, and $\mu_k$ is the $k$-th cluster centroid. (We use mini-batch k-means~\cite{batchkmeans}).
\item Use the cluster labels to train a pixel classifier using standard cross entropy loss.
\begin{align}
&\min_{\theta, \mathbf{w}} \sum_{i,p} \mathcal{L}_{CE}(g_\mathbf{w}(f_\theta(x_i)[ p]), y_{ip}) \\
&\mathcal{L}_{CE}(g_\mathbf{w}(f_\theta(x_i)[p]), y_{ip}) = - \log \frac{e^{s_{y_{ip}}}} {\sum_k e^{s_k}}
\end{align}
where $s_k$ is the $k$-th class score output by the classifier $g_\mathbf{w}(f_\theta(x_i, p))$.
\end{enumerate}

Given this baseline, we now propose the following modifications.

\subsection{Non-parametric prototype-based classifiers}
\label{nonparam}
The DeepCluster inspired framework above uses a separate, learned classifier.
However, in the unsupervised setting with constantly changing pseudo-labels, training a classifier jointly with the feature representation can be challenging.
An insufficiently trained classifier can feed noisy gradients into the feature extractor, resulting in noisy clusters for the next training round.

We therefore propose to jettison the parametric pixel classifier $g_\mathbf{w}$ entirely.
Instead, we label pixels based on their distance to the centroids (``prototypes''~\cite{protonet}) estimated by k-means.
This results in the following changed objective.
\begin{align}
&\min_{\theta} \sum_{i,p} \mathcal{L}_{clust}(f_\theta(x_i)[p], y_{ip}, \boldsymbol{\mu}) \\
&\mathcal{L}_{clust}(f_\theta(x_i)[p], y_{ip}, \boldsymbol{\mu}) =  -\log \Bigg( \frac{e^{-d(f_\theta(x_i)[p], \mu_{y_{ip}})}}{\sum_{l}e^{-d(f_\theta(x_i) [p], \mu_l)}} \Bigg)
\end{align}
where 
$d(\cdot, \cdot)$ is cosine distance.



\subsection{Invariance and Equivariance}
\label{inv_eqv}
Jointly learning the feature representation along with the clustering as above will certainly produce clusters that are compact in feature space, but there is no reason why these clusters must be semantic.
To get a semantic grouping of pixels, we need to introduce an additional inductive bias.
What must this inductive bias be if we have no labels?

The inductive bias we introduce is invariance to photometric transformations and equivariance to geometric transformations: the labeling should not change if the pixel colors are jittered slightly, and when the image is warped geometrically, the labeling should be warped similarly.
Concretely, if $Y$ is the output semantic labeling for an image $x$, and if $P$ and $G$ are photometric and geometric transformations respectively, then the output semantic labeling of a transformed image $G(P(x))$ should be $G(Y)$.

Implementing this constraint in a joint clustering and learning framework is tricky, since there isn't a ground truth label for each image.
The pseudo-ground truth labeling is itself derived from clustering, which is itself produced from the feature maps, and as such itself sensitive to input transformations.
Invariance/equivariance in this case therefore means two things: one, we should produce the same \emph{clusters} irrespective of the transformations, and two, the predicted \emph{pixel labels} should exhibit the desired in/equivariance.

\begin{figure}
\centering\noindent
\includegraphics[width=\linewidth]{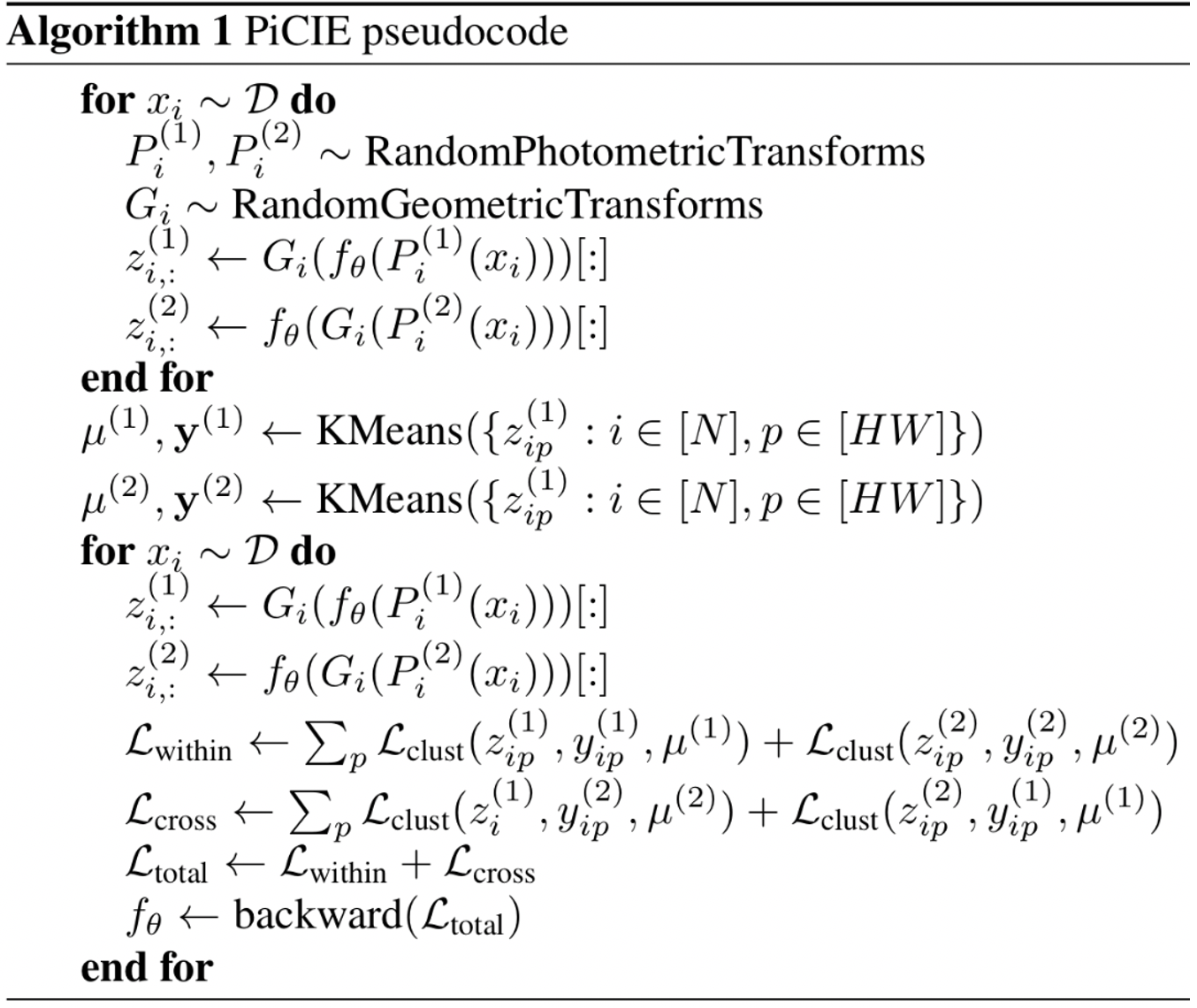}
\vspace{-8mm}
\caption*{\textbf{Above}: PiCIE pseudo-code. Notations consistent with Sec.~\ref{inv_eqv}. }
\label{fig:pseudocode}
\vspace{-4mm}
\end{figure}


 
\vspace{-2mm}
\subsubsection{Invariance to photometric transformations}
\vspace{-2mm}
We first address the question of invariance.
For each image $x_i$ in the dataset, we randomly sample two photometric transformations, $P_i^{(1)}$ and $P_i^{(2)}$. 
This yields two feature vectors for each pixel $p$ in each image $x_i$:
\begin{align}
z_{ip}^{(1)} = f_\theta(P_i^{(1)}(x_i))[p] \\
z_{ip}^{(2)} = f_\theta(P_i^{(2)}(x_i))[p]
\end{align}

We then perform clustering separately in the two ``views'' to get two sets of pseudo-labels and centroids:
\vspace{-2mm}
\begin{align}
\mathbf{y}^{(1)}, \boldsymbol{\mu}^{(1)} = \arg\min_{\mathbf{y}, \boldsymbol{\mu}} \sum_{i,p} \|z_{ip}^{(1)}  - \mu_{y_{ip}}\|^2 \\
\mathbf{y}^{(2)}, \boldsymbol{\mu}^{(2)} = \arg\min_{\mathbf{y}, \boldsymbol{\mu}} \sum_{i,p} \| z_{ip}^{(2)} - \mu_{y_{ip}}\|^2
\end{align}
\vspace{-4mm}

Given these two sets of centroids and these two sets of pseudo-labels, we use two sets of loss functions:
\begin{enumerate}
\item As before, we want the feature vectors to adhere to the clustering labels. Now that we have two views, we want this to be true in each view:
\begin{align}
\mathcal{L}_{within} = \sum_{i,p}&\mathcal{L}_{clust} (z_{ip}^{(1)}, y_{ip}^{(1)}, \boldsymbol{\mu}^{(1)}) \nonumber \\
&+  \mathcal{L}_{clust} (z_{ip}^{(2)}, y_{ip}^{(2)}, \boldsymbol{\mu}^{(2)})
\end{align}
\vspace{-8mm}

\item Because we posit that the clustering should be invariant to photometric transformations, we also want feature vectors from one view to match the cluster labels and centroids of the other:
\begin{align}
\mathcal{L}_{cross} = \sum_{i,p}&\mathcal{L}_{clust} (z_{ip}^{(1)}, y_{ip}^{(2)}, \boldsymbol{\mu}^{(2)})   \nonumber\\ 
&+ \mathcal{L}_{clust} (z_{ip}^{(2)}, y_{ip}^{(1)}, \boldsymbol{\mu}^{(1)})
\end{align}
\vspace{-8mm}

\end{enumerate}

This multi-view framework and the cross-view loss achieve two things. First, by forcing feature vectors from one transformation to adhere to labels produced by another, it encourages the network to learn feature representations that will be \emph{labeled} identically irrespective of any photometric transformations.
Second, by forcing the same feature representation to be consistent with two different clustering solutions, it encourages the two solutions themselves to match, thus ensuring that the set of concepts discovered by clustering is invariant to photometric transformations.

\vspace{-2mm}
\subsubsection{Equivariance to geometric transformations}
\vspace{-2mm}
A house and a zoomed-in version of the house should be labeled similarly, but may produce vastly different features. 
More precisely, the segmentation of the zoomed-in house should be a zoomed-in version of the original segmentation.
This is the notion of \emph{equivariance} 
to geometric transformations (such as random crops), which we add in next.

To learn equivariance with respect to geometric transformations, we sample a geometric transformation (concretely, random crop and horizontal flip) $G_i$ for each image. Then, in the above framework,  one view uses feature vectors of the transformed image, while the other uses the transformed feature vectors of the original:
\begin{align}
z_{ip}^{(1)} = f_\theta(G_i(P_i^{(1)}(x_i)))[p] \\
z_{ip}^{(2)} = G_i(f_\theta(P_i^{(2)}(x_i)))[p] 
\end{align}
The other steps are exactly the same. The two views are clustered separately, and the final training objective is the combination of the within-view and cross-view objectives:
\begin{align}
\mathcal{L}_{total} = \mathcal{L}_{within} + \mathcal{L}_{cross}
\end{align}
\vspace{-8mm}

\begin{figure*}
\centering
\includegraphics[width=0.95\linewidth]{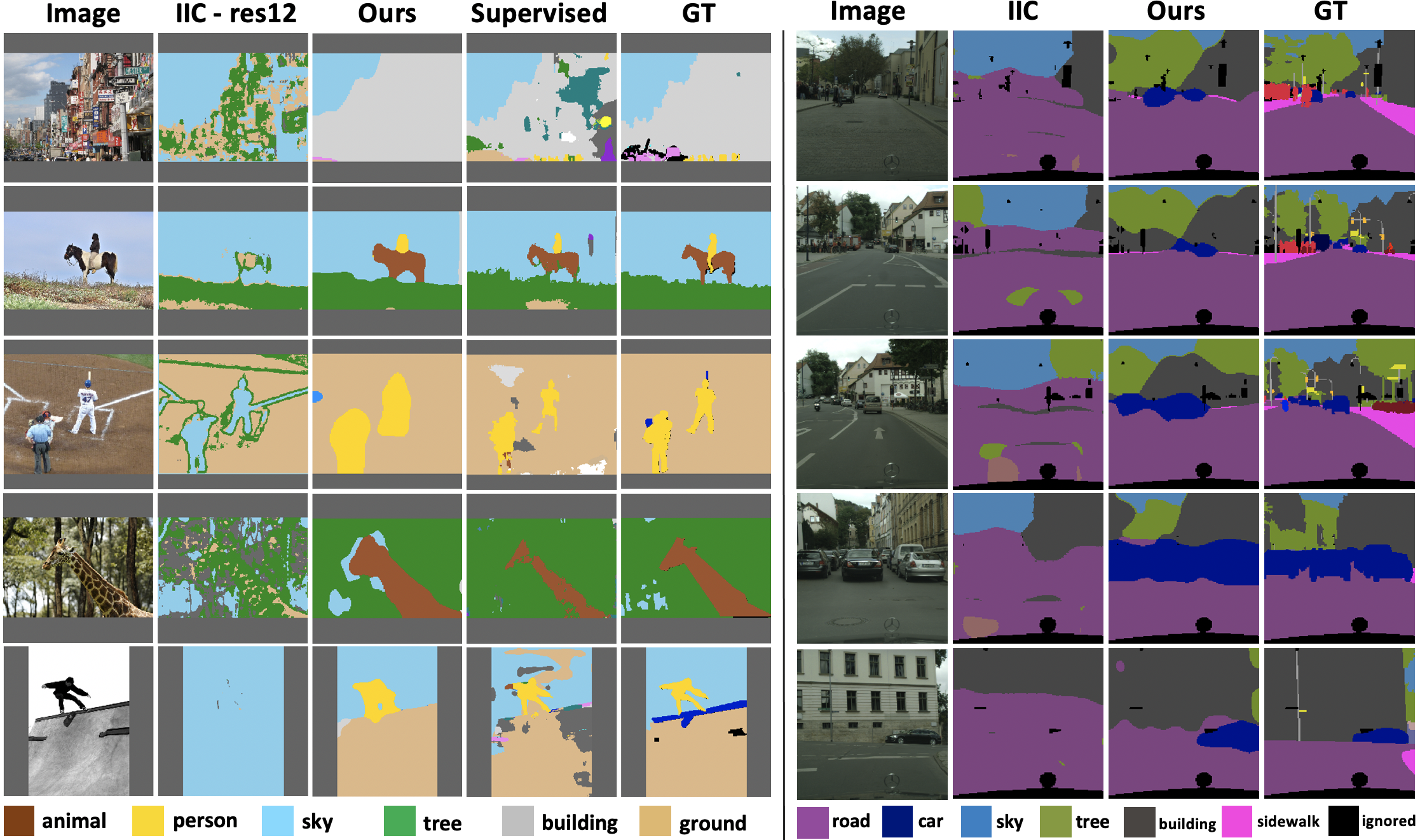}
\vspace{-2mm}
\captionof{figure}{Overall qualitative results on COCO-\emph{All}~\cite{coco} (\textbf{left}) and Cityscapes~\cite{cityscapes}(\textbf{right}). Note that we show IIC-res12 for COCO and IIC for Cityscapes to show the best result of the method on each dataset. Each ground truth label is assigned a color and for each cluster, the majority label's color is used. We show some of the color and name matches for better understanding. More in supplementary materials.  
}
\vspace{-4mm}
\label{fig:coco_vis}
\end{figure*}

\section{Experiments}

\subsection{Training details} For all our experiments, we use the Feature Pyramid Network~\cite{fpn} with ResNet-18~\cite{resnet} backbone pre-trained on ImageNet~\cite{imagenet}. The fusion dimension of the feature pyramid is 128 instead of 256. 
We apply L2 normalization on the feature map of our network. 
The cluster centroids are computed with mini-batch approximation with GPUs using the FAISS library~\cite{batchkmeans, faiss}. 
For the baselines, we do not use image gradients as an additional input when we use ImageNet-pretrained weight. Except in Table~\ref{tab:coco_scratch}, all images are resized and center-cropped to $320\times320$ during training. We used the published codes~\cite{dc1, iic} with minimal modification for the baselines. Other details are in supplementary.

\vspace{-2mm}
\paragraph{Pre-trained vs random initialization.} Prior works~\cite{iic, ac} train the network from random initialization, but for semantic segmentation it is unnecessary; unlike representation learning literature~\cite{moco, simclr, byol, localagg, dc1, dc2, odc}, our goal is to segment a given dataset as accurately as possible, and in a practical scenario one will always choose to initialize from a pre-trained network such as on the ImageNet dataset~\cite{imagenet}. Therefore, we train all models with ImageNet-pretrained weights, except that in Table~\ref{tab:coco_scratch} we show PiCIE outperforms all the baselines when trained from scratch as well. 

\vspace{-2mm}
\paragraph{Loss Balancing and Overclustering.} As shown in~\cite{dc1, dc2, iic}, jointly optimizing for a separate set of clusters with higher number improves the stability of clustering as well as the accuracy of the prediction. 
However, in unsupervised settings hyper-parameter tuning is often infeasible. Thus, we use the generic approach to balance the loss: 
\begin{align}
\mathcal{L} = \lambda_{K_1} \mathcal{L}_{K_1} + \lambda_{K_2} \mathcal{L}_{K_2}
\end{align}

$\lambda_{K_1} = \frac{\log K_2}{\log K_1 + \log K_2}$ and $\lambda_{K_2} = \frac{\log K_1}{\log K_1 + \log K_2}$ where $K_1$ and $K_2$ are the number of clusters. The intuition is that the magnitude of the cross-entropy loss depends logarithmically on the number of clusters, hence we prevent the overclustering to overwhelm the gradient flow. We fix $K_2=100$ and add ``+H.'' in results when applied. Similarly, due to the imbalance of datasets, the computed clusters will have largely different sizes; we apply a  balance term for each cluster during the cross-entropy computation.


\subsection{Baselines}
We describe the baseline methods that we compare PiCIE to: IIC~\cite{iic} and modified DeepCluster~\cite{dc1} for segmentation purposes. They are state-of-the-art \emph{implicit} and \emph{explicit} clustering-based learning methods. 

\vspace{-2mm}
\paragraph{IIC.}
IIC~\cite{iic} is an implicit clustering method where the network directly predicts the (soft) clustering assignment of each pixel-level feature vector. The main objective is maximizing the mutual information between the predictions of a pixel and neighboring pixel(s). For controlled experiments, we use FPN with ResNet-18 same as PiCIE as well as the first two residual blocks of ResNet-18 (IIC -- \emph{res12}) similar to the original shallow VGG-like~\cite{vgg} model (details in supplementary). Following the original paper~\cite{iic}, we used auxiliary over-clustering loss with $K=45$.

\vspace{-2mm}
\paragraph{Modified DeepCluster.} DeepCluster is an explicit clustering method where the network clusters the feature vectors of given images and uses the cluster assignment as labels to train the network. To adjust to our problem setup, we modify the original DeepCluster to instead cluster pixel-level feature vectors before the final pooling layer. This allows the network to assign a label to each pixel. However, since the size of image explodes the number of feature vectors to cluster, we apply mini-batch k-means~\cite{batchkmeans} to first compute the cluster centroids, assign labels, and train the network.

\begin{figure*}
\centering\noindent
\includegraphics[width=\linewidth]{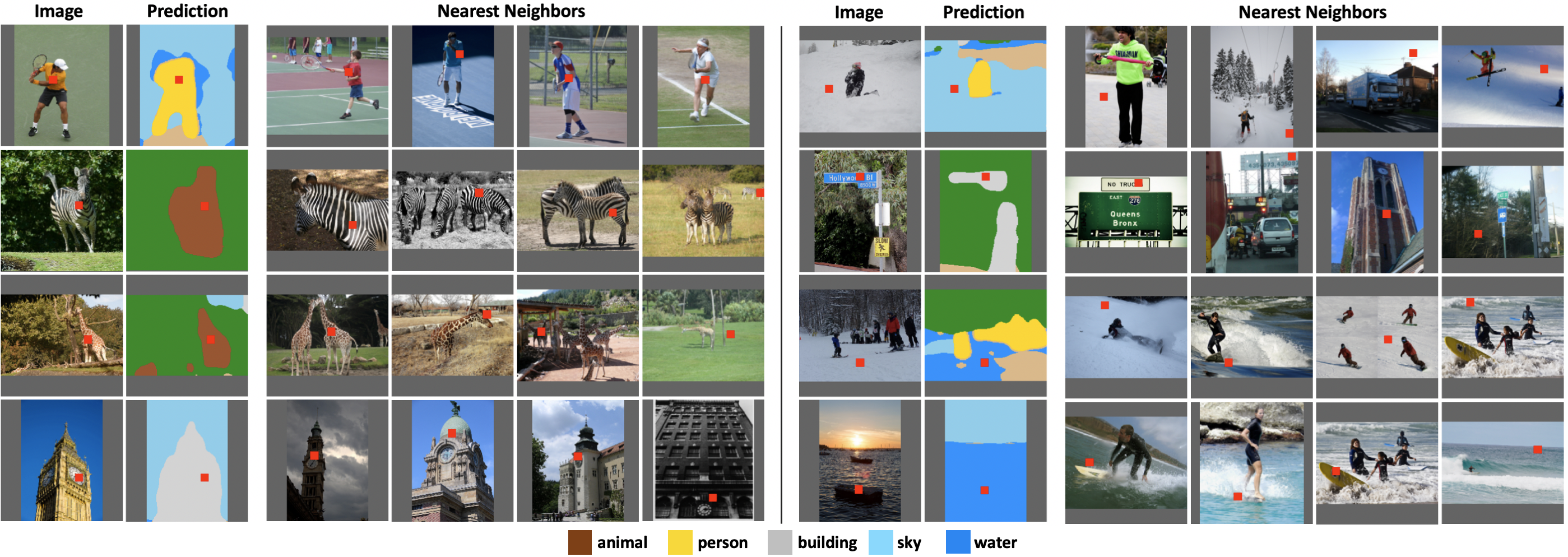}
\vspace{-6mm}
\caption{Nearest neighbor results for correctly predicted (\textbf{left}) instances and incorrectly predicted (\textbf{right}) instances. The red box indicates the position of the particular feature vector (size exaggerated). More details in supplementary materials. }
\vspace{-4mm}
\label{fig:nearest_neighbors}
\end{figure*}


\subsection{Datasets}

\vspace{-2mm}
\paragraph{COCO.} Following \cite{iic}, we evaluate our model on the COCO-Stuff dataset~\cite{cocostuff}. The COCO-Stuff dataset is a large-scale scene-centric dataset of images with 80 \emph{things} categories and 91 \emph{stuff} categories. We follow the same preprocess as~\cite{iic} where classes are merged to form 27 (15 \emph{stuff} and 12 \emph{things}) categories. 
Unless otherwise stated, \textbf{we evaluate both \emph{things} and \emph{stuff} categories}, unlike prior works which evaluate only \emph{stuff}.



\vspace{-2mm}
\paragraph{Cityscapes.}
We further evaluate our model on the Cityscapes dataset~\cite{cityscapes}. Cityscapes is a set of images of street scenes from 50 different cities. 
There are 30 classes of instances that can be further categorized into 8 groups. After filtering out \emph{void} group, we have 27 categories. We train our method as well as IIC and modified DeepCluster with $K=27$ where $K$ is the number of clusters. 

\subsection{Results}
\input{main_results_table.tex}

\input{coco-partition-table.tex}

\input{cityscapes_table.tex}

\input{coco_stuff_from_scratch_table.tex}

In Table~\ref{tab:coco_main}, we compare PiCIE with the following baselines: \emph{No Train}, \emph{modified DeepCluster}~\cite{dc1}, and \emph{IIC}~\cite{iic}. Unlike the prior works~\cite{iic, ac} where only \emph{stuff} categories are considered, we evaluate the models on both \emph{stuff} and \emph{things} categories to test on more realistic setting. Since the majority of scene-centric image dataset consists of \emph{stuff} categories, our evaluation now faces a severe imbalance problem. Also, the learning mechanism of IIC assumes local spatial consistency, which is not often true for \emph{things} categories due to more dynamic shape variations. We found that IIC tends to overfit to low-level visual cues since (implicit) clustering is done within a batch and insufficient supervisory signal is present when an instance has dynamic and complex visual cues. Indeed, in Figure~\ref{fig:coco_vis} no \emph{things} categories are correctly segmented from IIC results. On the other hand, PiCIE's novel in/equivariance loss enforces geometric consistency as an inductive bias to learn high-level visual concepts, and as shown in Figure~\ref{fig:coco_vis} PiCIE (``Ours'') is capable of segmenting both \emph{stuff} and \emph{things} categories with high accuracy. As a result, Table~\ref{tab:coco_main} shows that PiCIE largely outperforms other baselines (+ \textbf{17.5} Acc. and \textbf{4.5} mIoU). In Table~\ref{tab:cityscapes}, we test the baselines and our method on Cityscapes and show similar level of advantages (+ \textbf{18} Acc. and \textbf{5.3} mIoU). Finally, Table~\ref{tab:coco_scratch} shows PiCIE outperforms the other models on the benchmark from~\cite{iic, ac} where the image size is 128$\times$128, models are trained from scratch, and only \emph{stuff} labels are considered for evaluation. 



\vspace{-2mm}
\paragraph{Things vs stuff. } In Table~\ref{tab:coco_partitions}, we show that PiCIE improves mainly on \emph{things} categories (+\textbf{10} mIoU) while maintaining better or compatible performance on \emph{stuff} categories compared to other methods. This indicates that enforcing geometric transformation equivariance was highly effective on \emph{things} categories where the instances objects with distinct shape and boundaries. Furthermore, we show in Table~\ref{tab:coco_partitions} and~\ref{tab:coco_scratch} that PiCIE still outperforms on \emph{stuff} categories with or without ImageNet-pretrained weights.


\subsection{Ablation Study}
\input{ablation_1_table.tex}

\input{ablation_2_table.tex}

In Table~\ref{tab:coco_ablation1}, we decompose our method to examine which component affects the performance the most. We gain 5 points by using a non-parametric classifier with cluster centroids. We further gain 3 points with cross-view learning with invariance transformations. Equivariance learning adds another 5.5 points, and with auxiliary over-clustering, we arrive at 49.99 pixel accuracy and 14.36 mIoU. 

In Table~\ref{tab:coco_ablation2}, we test alternatives of different components of PiCIE. 
First, one could wonder if our cross-view loss can be replaced by MSE loss, directly minimizing the feature vectors of the two views. This leads PiCIE to a suboptimal solution: 1) the direct distance between two feature vectors can be overwhelmed by low-level or irrelevant signals whereas cross-view loss directs the gradient to the nearest centroid, hence only considers relevant signals and 2) MSE loss requires hyperparameter tuning to be jointly used with cross-entropy loss, which is infeasible in purely unsupervised setting. 
Also, one could doubt if two sets of clustering are necessary; a single clustering with geometric transformation on the predicted labels can be used as an alternative to compute the cross-view loss. However, the two versions of an image contain different information (e.g., \emph{zoomed-in} vs \emph{full house}) that can be mutually beneficial. We test them all (and more) in Table~\ref{tab:coco_ablation2} and the results justify our choices.

\subsection{Analysis}



\vspace{-2mm}
\paragraph{Nearest neighbor analysis.} In Figure~\ref{fig:nearest_neighbors}, we show the nearest neighbors of correctly (left) and incorrectly (right) predicted instances. The nearest neighbors of correctly predicted segments share close high-level semantics (e.g., \emph{person playing tennis, zebra, giraffe}, and \emph{a building with a clock}). This indicates that intra-class semantics are well preserved. 
The incorrectly predicted segments also have semantically and visually close nearest neighbors. For example, the first row shows that \emph{snow} pixels are confused with \emph{sky} as the two concepts are visually alike. Such visual ambiguity is an inherent limitation of unsupervised methods.

\vspace{-2mm}
\paragraph{Representation quality. } In Table~\ref{tab:coco_linear}, we compare the learned representations by training a linear classifier for each trained method from our main experiments on COCO-\emph{All}. We train with $\eta=0.001$ for 10 epochs with cross-entropy loss. This allows us to analyze whether the difficulty is from the representation or from clustering. Compared to the unsupervised results from Table~\ref{tab:coco_main}, baselines have a huge performance gap whereas PiCIE has a minimal gap. This indicates that clustering is where the major difficulty is and PiCIE gives close-to-optimal clustering given learned representation. In Table~\ref{tab:coco_finetune}, we show that PiCIE can give better network initialization for supervised training.

\input{transfer_learning_table.tex}

\input{retraining_table.tex}

\vspace{-2mm}
\section{Conclusion}
In this paper, we introduced a new framework for unsupervised semantic segmentation with clustering. Our main contribution is to incorporate geometric consistency as an inductive bias to learn invariance and equivariance for photometric and geometric variations. Our novel cross-view loss is simple yet highly effective in learning high-level visual concepts necessary to segment \emph{things} categories. Our method is the first unsupervised semantic segmentation that works for both \emph{stuff} and \emph{things} categories without rigorous hyper-parameter tuning or task-specific pre-processing.

\section*{Acknowledgements} 
This work was supported by DARPA Learning with Less Labels program (HR001118S0044), CHS-1617861, CHS-1513967, CHS-1900783, and CHS-1930755. 

\clearpage
{\small
\bibliographystyle{ieee_fullname}
\bibliography{main}
}
\end{document}


\title{PiCIE: Unsupervised Semantic Segmentation using Invariance and Equivariance in Clustering -- \textit{Supplementary Materials}}

\author{Jang Hyun Cho$^1$
\quad\quad\quad
Utkarsh Mall$^2$
\quad\quad\quad
Kavita Bala$^2$
\quad\quad\quad
Bharath Hariharan$^2$\\
$^1$University of Texas at Austin\quad\quad\quad\quad $^2$Cornell University\\
}

\maketitle

\section*{S1. More experiment details}

\subsection*{Network architectures}
For every method, we used the Feature Pyramid Network~\cite{fpn} to effectively encode representations from multiple scales. However, we only use pixel-wise randomly initialized linear ($1\times1$ convolutional) layer for each level of the intermediate feature maps from ResNet-18~\cite{resnet}. As noted in the main paper, we projected each of the feature maps to $128$ dimensions instead of $256$ from the FPN. After the linear projection, we directly bilinear-upsampled to $1/4$ scale of image resolution and element-wise summed to get the final $128\times H\times W$ representation without the last $3\times3$ convolutional \textit{smoothing} layers ($H=W=80$ during training with $320\times320$ images). Note that this is a simplified version of the semantic segmentation branch of Panoptic FPN~\cite{panoptic_fpn}, a simple application of FPN to segmentation task. At the end, the only added parameters from ResNet-18 are 4 $1\times1$ convolutional layers.

\paragraph{IIC}
For controlled experiments, we changed the network architecture of default IIC from the original shallow VGG-like model to FPN with ResNet-18 as described above. Following the original paper~\cite{iic}, we used auxiliary over-clustering loss: We kept the original $k=45$ since the difference was minimal between $k \in \{45, 100, 250 \}$. Also, the original IIC objective has a hyper-parameter $\lambda$ which controls the ``strictness'' of the uniform distribution of clustering constraint. This could potentially alleviate the problem that IIC faces. In Table~\ref{tab:iic_lambda} we tested with $\lambda \in \{ 1, 1.25, 1.5, 1.75, 2, 3 \}$ on COCO-\textit{All} and $\lambda=1$ performed the best, hence we kept $\lambda=1$ in all our experiments. Similarly, we tested different learning rates $\eta \in \{ 0.1, 0.01, 0.001, 0.0001 \}$ and $\eta=0.0001$ was optimal. Both of these $\lambda$ and $\eta$ coincide with those in the original paper.

\paragraph{IIC-res12.} We discovered that the shallow version of IIC performs better qualitatively on the (processed) COCO dataset~\cite{iic}. This is because a shallow network tends to overfit to low-level visual signals such as color and texture due to its narrow receptive field. Since the dataset is pre-processed to reduce images that have too many pixels in the \textit{things} categories, which are often visually more complex, perhaps the shallow IIC can be more effective for solving simple background segmentation compared to deep IIC. Therefore, we tested both versions. Note that the shallow VGG-like network used in the original IIC paper is unable to load ImageNet-pretrained weight, hence we instead used the first two residual layers \textit{res1} and \textit{res2} of ResNet-18~\cite{resnet} as an alternative. They have nearly the same number of parameters and in the main paper, Table 3, we show that IIC with \textit{res12} achieves similar accuracy on the original COCO-\textit{Stuff} benchmark (27.7 and 27.92)~\cite{iic}. Similar to IIC, we apply auxiliary over-clustering with $k=45$.

\paragraph{Modified DC.} Since DeepCluster~\cite{dc} was originally designed for the task of image clustering, we modified the framework to fit the task of segmentation (pixel-wise classification). The network alternates between computing pseudo-labels and training. As mentioned in the main paper, the representation is pixel-level by removing the final pooling layer. This makes storing the feature vectors of the entire dataset infeasible, so we perform mini-batch k-means to first estimate cluster centroids, assign pseudo-labels, and train the network with the pseudo-labels. The same set of transformations as PiCIE is used on each image during training. Note that similar to IIC, image gradient is not concatenated in the input when initialized from ImageNet-weight. We do not apply over-clustering since the model without over-clustering performed the best compared to $k\in \{100 ,250, 1000, 2500 \}$.

\subsection*{Datasets}
For training modified DC and PiCIE, we used simple pre-processing: resizing and center-crop to $320\times320$. For IIC, we used the original paper's pre-processing with their published code. 

\paragraph{Transformations.} For photometric transformations, we randomly applied \textit{color jitter}, \textit{gray scale}, and \textit{Gaussian blur}. Random jitter  consists of jittering brightness, contrast, saturation, and hue. All jittering transformations are applied with probability $p=0.8$ and control factors $0.3, 0.3, 0.3, 0.1$, respectively. Random gray scale is applied with probability of $p=0.2$. Random Gaussian blur is applied with probability of $p=0.5$ and radius randomly chosen: $\sigma \in [0.1, 2]$. For geometric transformations, we applied \textit{random crop} and \textit{random horizontal flip} with crop factor $r\in[0.5, 1]$ and flipping probability $p=0.5$. In order to ensure that the same transformations are applied during clustering and training, we first sample transformations during clustering and store them in a list to re-use during training.These hyper-parameters are a standard choice adopted in many other works~\cite{moco, simclr, simclrv2}.

\input{iic_lambda.tex}

\subsection*{Training}

\paragraph{Clustering.} The cluster centroids are computed with mini-batch k-means with GPUs using the FAISS library~\cite{faiss}. The initial cluster centroids are computed with 50 batches with batch size of 128, then the centroids are updated every 20 iterations. For every other hyperparameters related to clustering, we followed Caron et al.~\cite{dc}. Since this process is highly optimized, it takes about 20 minutes to prepare the pseudo-labels for training every epoch on the COCO dataset, which makes less than half for training the network in total compared to IIC using the published code. 

\paragraph{Training details.}
We trained every method with 10 epochs when trained with ImageNet weight initialization, and 20 epochs when trained from scratch. For modified DC and PiCIE, we used ADAM optimizer with learning rate $\eta = 1\times 10^{-3}$, $\beta=(0.9, 0.999)$ and weight decay 0. For IIC, their original hyperparameter setting was better, so we kept their setting ($\eta=1\times 10^{-4})$. For the transfer learning and supervised training experiments, we used $\eta = 1\times 10^{-3}$, $\beta=(0.9, 0.999)$ and weight decay 0, consistent with the setting from the main experiments. For the final objective, we applied weighted cross-entropy loss with per-cluster weight is balanced with the size of each cluster. We simply average the \textit{cross} and \textit{within} losses.  

\paragraph{Evaluation metric.} For evaluating our model, we followed the evaluation metric from~\cite{iic} with pixel accuracy after Hungarian-matching~\cite{hungarian} the cluster assignments to the ground truth labels. We also report mean IoU to account for false positives and negatives. In Table 2 of the main paper, we compute the accuracy and mIoU from the same model trained on COCO-\textit{All} ($K=27$), but evaluated by only accounting for the labels in each partition. This can be done efficiently by computing the confusion matrix of the all classes $K=27$ first and partitioning the matrix accordingly. In Table 3 of the main paper, we closely follow the experiment setting of ~\cite{iic}: the image resolution is $128\times128$, the images are pre-scaled and constant-padded, and $K=15$ which means only \textit{stuff} categories are considered for evaluation.

\subsection*{Visualizations}
For producing consistent visualizations, we used majority vote for each obtained cluster. That is, we first assigned color values to each ground truth label and for each obtained set of clusters, we assign the color of the majority class. In the main paper, notice that we showed IIC-res12 for COCO and IIC for Cityscapes. We included the version that had better qualitative results. We hypothesize that since COCO was preprocessed to include more \textit{stuff} categories, it is easier for the shallow network which overfits to low-level cues (e.g., color and texture) to segment images well since the majority of \textit{stuff} instances are visually simple. For the nearest neighbor result, we first chose successful and failure results from the large set of randomly selected images (results below), picked a pixel coordinate of interest, and computed the nearest neighbor on the entire validation set of COCO-\textit{All}. Then, we extracted the images that the neighbors belong to and visualized.

\section*{S2. More results}
In this section, we show more qualitative results randomly chosen for both IIC and IIC-res12, as well as modified DC and PiCIE. 

\subsection*{Robustness on Color and Geometric transformations}
We show that PiCIE successfully learns photometric invariance and geometric equivariance by evaluating our model with test-time augmentation. We apply the same set of photometric transformations (color jitter, Gaussian blur, and greyscale) and geometric transformations (horizontal flip and random crop) and report the results in Table~\ref{tab:robust}. 


\input{robust.tex}

\section*{S3. Analysis}
We discuss a few possible directions for future study. Note that \textit{MDC} stands for modified DeepCluster. 

\paragraph{Visual ambiguity.} 
As shown in visualization, visual ambiguity leads to mis-classification of certain classes. Snowy ground is often confused with either sky or water, and grass on a flat ground is confused with ground. The core problem is twofold: First, the classification of the segment masks are done with cluster centroids, which follow the ``majority trend.'' For example, the majority of ``ground'' instances is not covered by snow, making the confidence low. Second, the visual similarity does not always correlate to the semantic similarity, and such discrepancy leads to confusion. ``Snow ground'' is often texture-less and mono-colore, similar to ``sky'' or ''water.'' This is an inherent limitation of unsupervised learning methods. 

\paragraph{Co-occurrence.}
Some foreground classes such as ``boat'' or ``airplane'', only occur surrounded by ``water'' or ``sky.'' Since \textit{stuff} categories have far more pixels, they are often \textit{subsumed} in the co-occurring background classes. We hypothesize that this effect will be mitigated if the dataset had more images of stand-alone ``boat'' or ``airplane.'' or with an effective way to contrast between the two entities such as using either a generic or a learned boundary detector, which can be a future work. 

\paragraph{Boundary precision.}
Since we do not have any supervision to train for precise boundaries, many foreground instances are segmented with over-confidence. Pixels around boundaries are hard samples to correctly predict. Using a generic edge detector or post-processing through iterative refinement such as CRF~\cite{crf} may improve the result, which is outside the scope of our project.

\begin{figure*}
\centering\noindent
\includegraphics[width=0.9\linewidth]{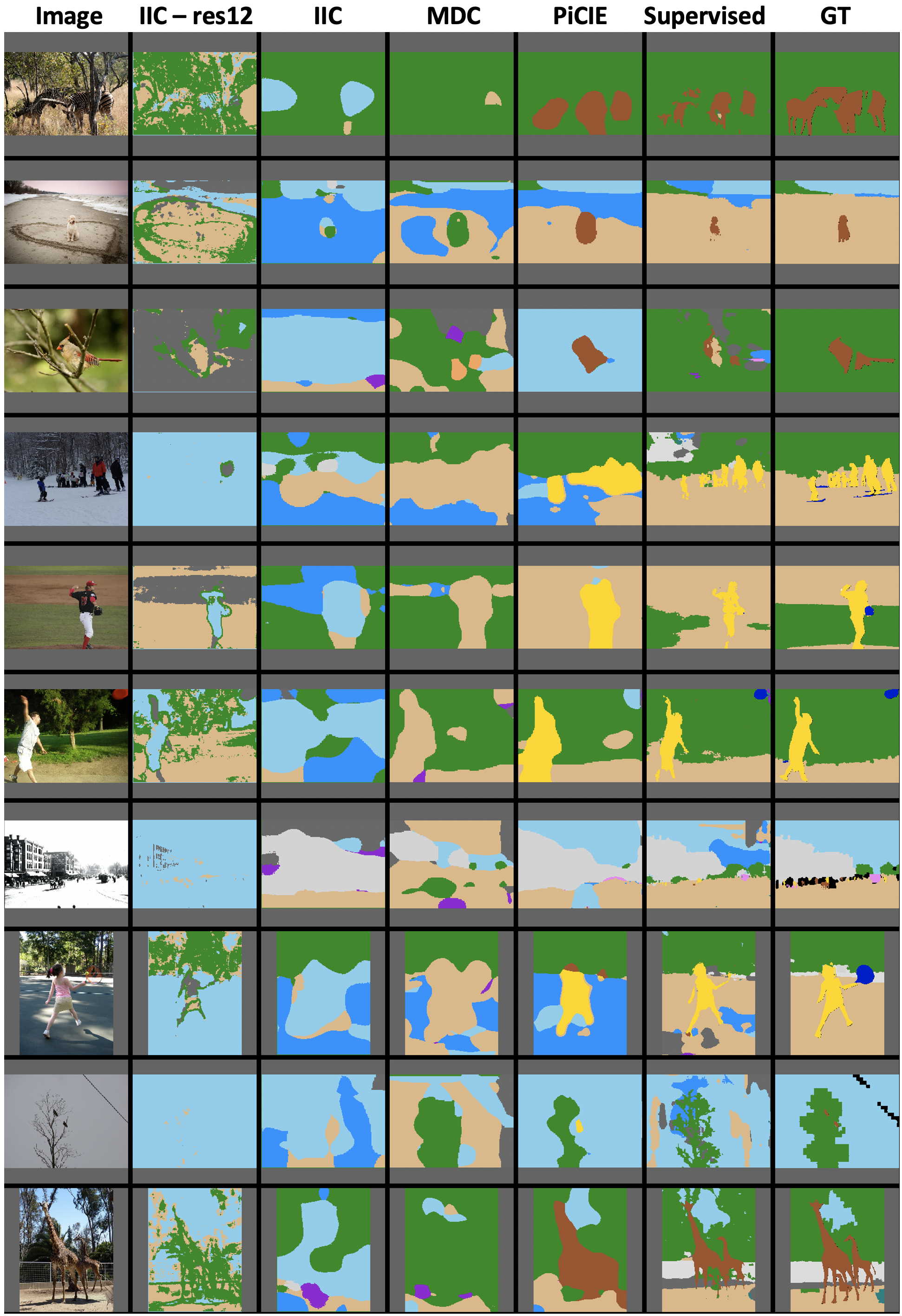}
\label{fig:coco1}
\end{figure*}


\begin{figure*}
\centering\noindent
\includegraphics[width=0.9\linewidth]{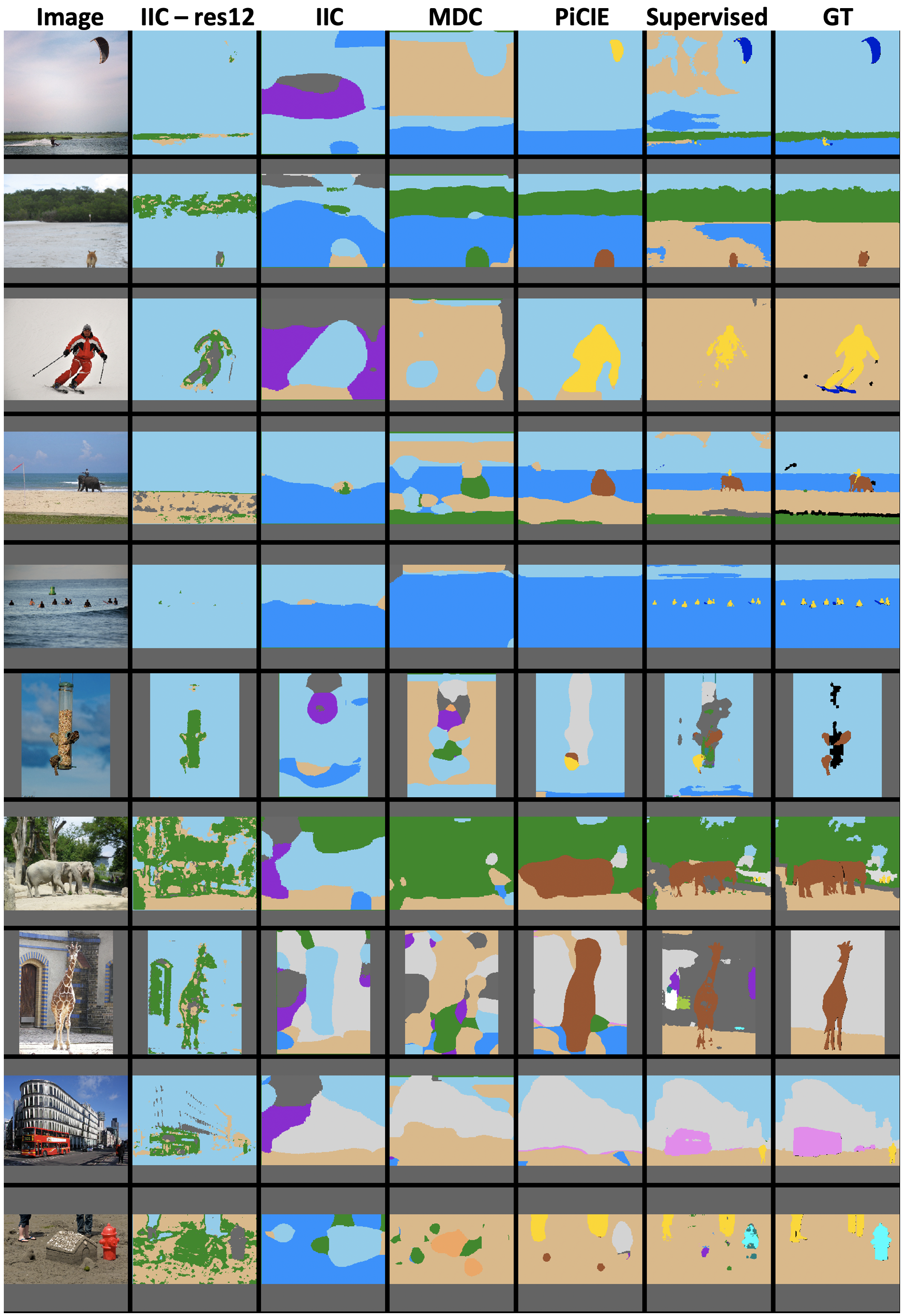}
\label{fig:coco1}
\end{figure*}

\begin{figure*}
\centering\noindent
\includegraphics[width=0.9\linewidth]{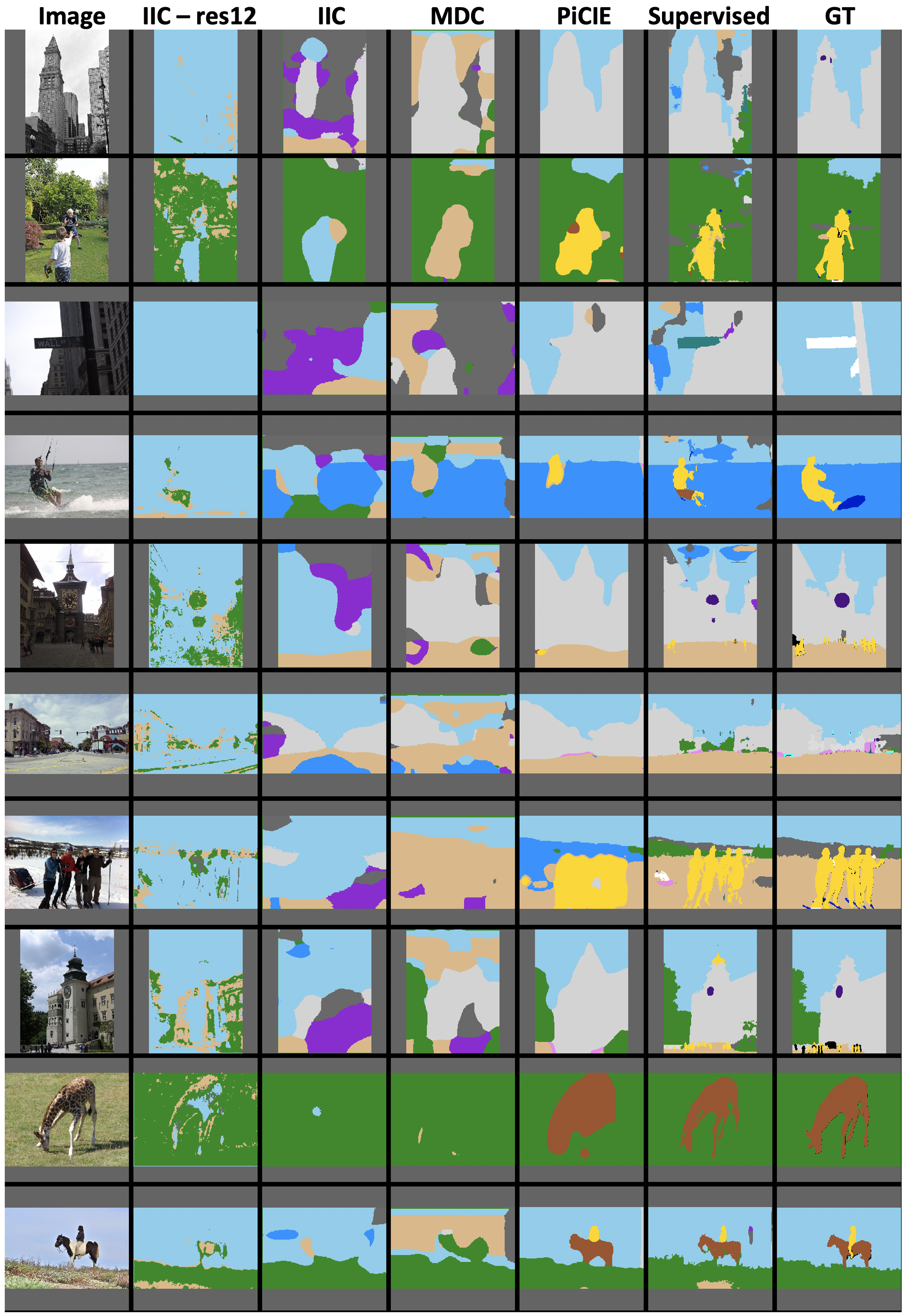}
\label{fig:coco1}
\end{figure*}

\begin{figure*}
\centering\noindent
\includegraphics[width=0.9\linewidth]{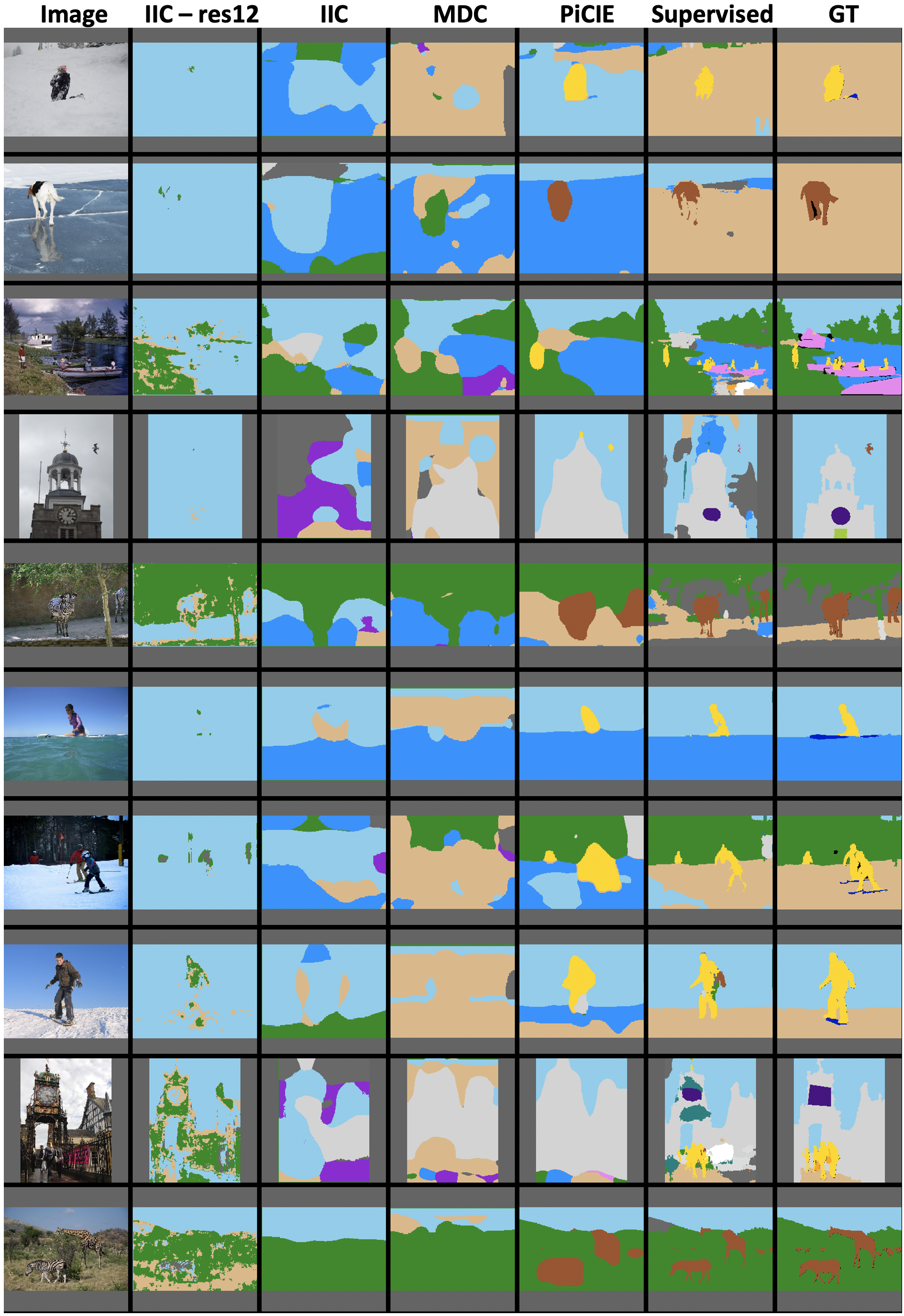}
\label{fig:coco1}
\end{figure*}



\begin{figure*}
\centering\noindent
\includegraphics[width=0.9\linewidth]{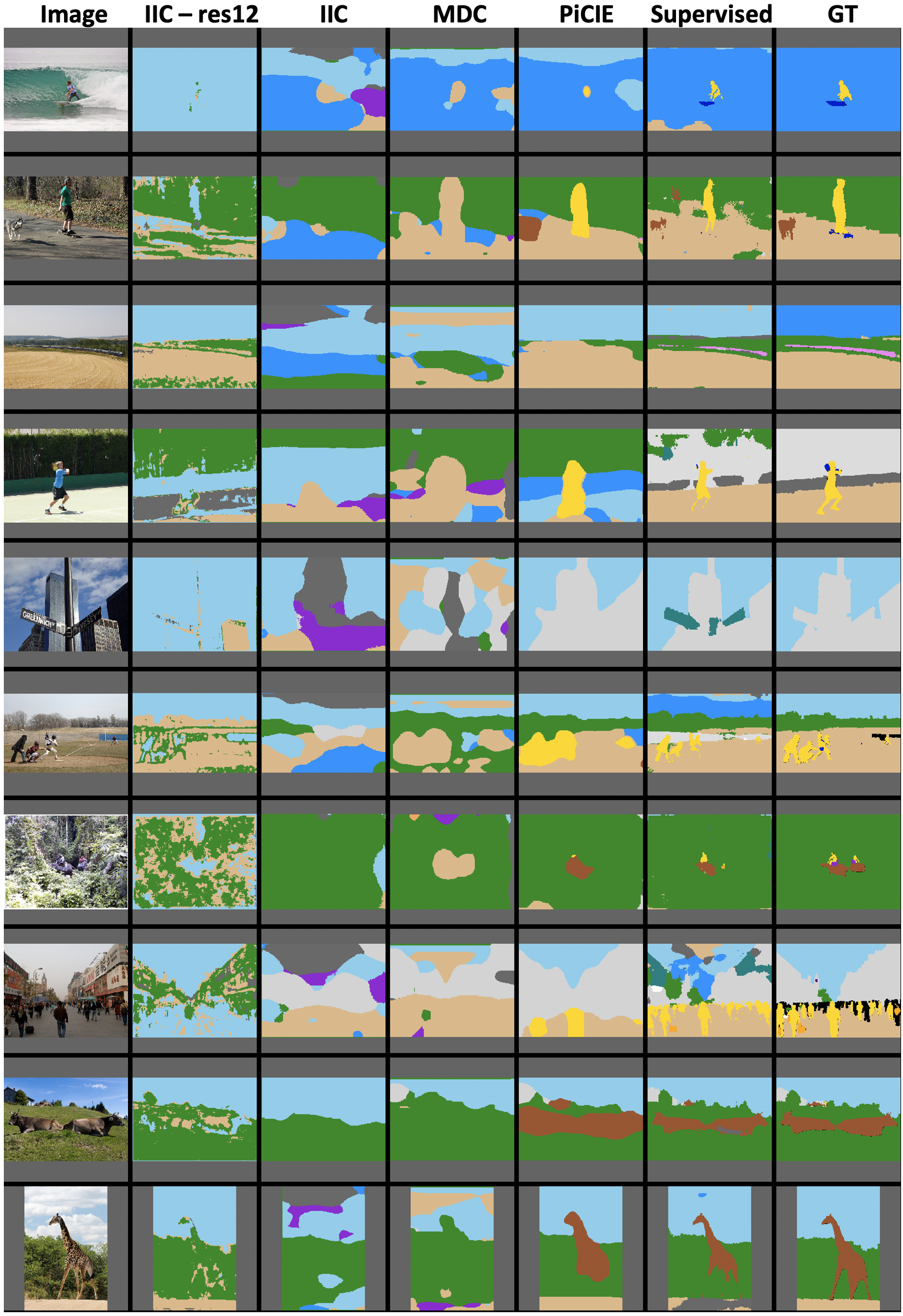}
\label{fig:coco1}
\end{figure*}


\begin{figure*}
\centering\noindent
\includegraphics[width=0.9\linewidth]{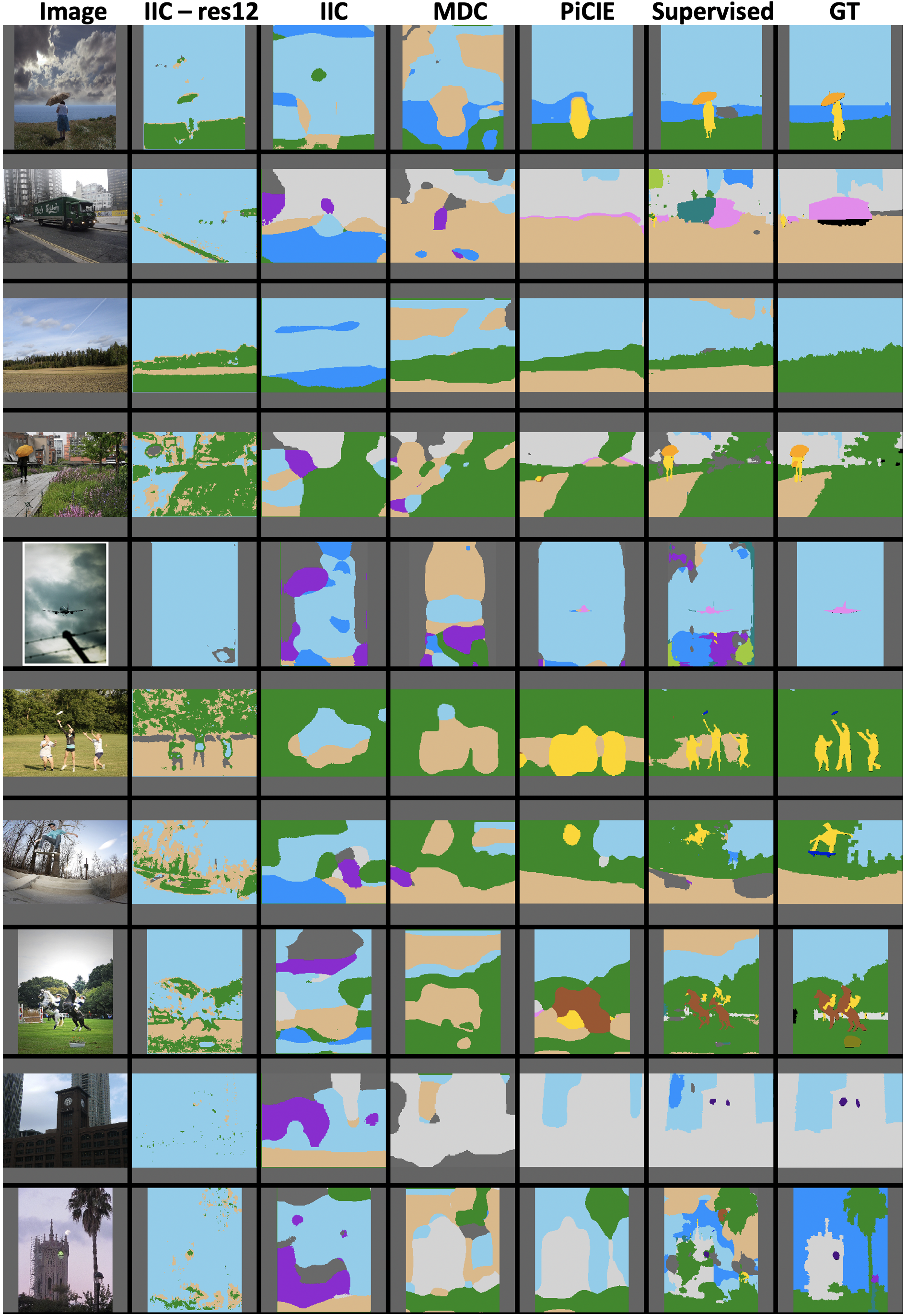}
\label{fig:coco1}
\end{figure*}

{\small
\bibliographystyle{ieee_fullname}
\bibliography{supplement}
}

%% file: main_results_table.tex
\begin{table}
\begin{center}
\resizebox{0.7\linewidth}{!}{
\begin{tabular}{llccc}
Method  & Classifier &   Acc. & mIoU \\
\toprule
No Train  & Linear &  17.45 & 3.70   \\
No Train  & Prototype &  26.26& 8.41 \\
Modified DC  & Linear &  32.21 & 9.79 \\
IIC - res12~\cite{iic}   & Linear &  22.45 & 4.11\\ 
IIC~\cite{iic}   & Linear &  21.79 & 6.71\\
\midrule
PiCIE  & Prototype &  48.09  & 13.84\\
PiCIE + H.  & Prototype &  \textbf{49.99} & \textbf{14.36} \\ 
\bottomrule
\end{tabular}
}
\end{center}
\vspace{-2mm}
\caption{\textbf{COCO-\emph{All}~\cite{iic} results.} Our method is compared to clustering methods adapted to semantic segmentation. ``+H.'' denotes PiCIE trained with auxiliary clustering.
}
\vspace{-8mm}
\label{tab:coco_main}
\end{table}

%% file: coco-partition-table.tex
\begin{table}
\begin{center}
\resizebox{0.85\linewidth}{!}{
\begin{tabular}{lcccc}
Method & Partition & \# Classes & Acc. & mIoU \\
\toprule
Modified DC~\cite{dc1} &  & &44.28 & \textbf{22.24} \\
IIC~\cite{iic} & \emph{Stuff} &15& 33.91 & 12.00\\
PiCIE + H. && &\textbf{74.56} & 17.32\\
\midrule
Modified DC~\cite{dc1} &  & &67.06&11.55\\
IIC~\cite{iic} & \emph{Things} &12&43.93 & 13.64 \\
PiCIE + H. & & & \textbf{69.39} &\textbf{23.83} \\
\midrule
Modified DC~\cite{dc1} &&& 32.21&9.79\\
IIC~\cite{iic} & \emph{All}&27& 21.79 & 6.71\\
PiCIE + H. && &\textbf{49.99}&\textbf{14.36}\\
\bottomrule
\end{tabular}
}
\end{center}
\vspace{-2mm}
\caption{Results on different partitions of the COCO dataset.}
\vspace{-6mm}
\label{tab:coco_partitions}
\end{table}

%% file: cityscapes_table.tex
\begin{table}
\begin{center}
\resizebox{0.7\linewidth}{!}{
\begin{tabular}{lccc}
Method & \# Classes & Accuracy & mIoU   \\
\toprule
IIC  &  \multirow{4}{*}{27}& 47.88 & 6.35\\
IIC -- res12  & & 29.78 & 4.96\\
Modified DC & &40.67 & 7.06\\ 
PiCIE & & \textbf{65.50} & \textbf{12.31}\\
\bottomrule
\end{tabular}
}
\end{center}
\vspace{-2mm}
\caption{Cityscapes results. }
\vspace{-6mm}
\label{tab:cityscapes}
\end{table}

%% file: coco_stuff_from_scratch_table.tex
\begin{table}
\begin{center}
\resizebox{0.5\linewidth}{!}{
\begin{tabular}{lc}
Method  & COCO-\emph{Stuff}   \\
\toprule
Random CNN & 19.4\\
K-means~\cite{km} & 14.1\\
SIFT~\cite{sift}& 20.2\\
Doersch 2015~\cite{doersch} & 23.1\\ 
Isola 2016~\cite{isola} & 24.3\\
DeepCluster~\cite{dc1} & 19.9 \\
IIC~\cite{iic}  & 27.7  \\
AC~\cite{ac} & 30.8 \\
\midrule
Modified DC & 25.26\\ 
IIC & 27.97\\
IIC -- res12 & 27.92\\
\midrule
PiCIE      &  \textbf{31.48}\\
\bottomrule
\end{tabular}
}
\end{center}
\vspace{-2mm}
\caption{COCO-\emph{Stuff} results without ImageNet pretrained weight following~\cite{iic, ac}. First section is from prior works~\cite{iic, ac} and the last two sections are from our implementation.  }
\vspace{-10mm}
\label{tab:coco_scratch}
\end{table}


%% file: ablation_1_table.tex
\begin{table}
\begin{center}
\resizebox{\linewidth}{!}{
\begin{tabular}{cccccc}
Nonpara- & Photo- & Geo- & Over- &  \multirow{2}{*}{Accuracy} & \multirow{2}{*}{mIoU} \\
metric & metric & metric & cluster & & \\
\toprule
&&&& 34.35&9.88\\ 
\checkmark&&&& 39.25&9.82\\
\checkmark&\checkmark&&& 42.55&9.84\\
\checkmark& & \checkmark&&46.97 & 12.04\\
\checkmark&\checkmark&\checkmark&&48.09 & 13.84\\
\checkmark&\checkmark&\checkmark&\checkmark&\textbf{49.99} & \textbf{14.36}\\
\bottomrule
\end{tabular}
}%
\end{center}
\vspace{-2mm}
\caption{\textbf{Ablation study 1.} Our method is decomposed to examine which components affect the performance the most.   }
\vspace{-6mm}
\label{tab:coco_ablation1}
\end{table}

%% file: ablation_2_table.tex
\begin{table}
\begin{center}
\resizebox{\linewidth}{!}{
\begin{tabular}{cccccc}
Single & MSE eqv.  &  No inv. & No balance &  Accuracy&mIoU \\
\toprule
&&&& \textbf{48.09}&\textbf{13.84}\\ 
&&&\checkmark& 40.56& 11.46\\
\checkmark&&&&44.31&11.71\\
&\checkmark&&& 44.15&10.98\\
\checkmark&&\checkmark&&41.70&9.92\\
\bottomrule
\end{tabular}
}
\end{center}
\vspace{-2mm}
\caption{\textbf{Ablation study 2.} One or more components in our method is replaced with alternative options. 
}
\vspace{-8mm}
\label{tab:coco_ablation2}
\end{table}

%% file: transfer_learning_table.tex
\begin{table}
\begin{center}
\resizebox{0.8\linewidth}{!}{
\begin{tabular}{lccc}
Feature Extractor  & Normalization &  Acc. & mIoU  \\
\toprule
Modified DC  &  & 50.79 & 13.76 \\ 
Modified DC  & \checkmark & 48.61 & 13.30  \\
IIC  && 51.49 &13.26   \\ 
IIC  & \checkmark& 44.50 & 8.37 \\
\midrule
No Eqv.  & &47.73 &12.59\\
No Eqv.  & \checkmark &48.58 &10.40\\ 
Single Cluster  & & 50.34 & 12.70 \\ 
Single Cluster  & \checkmark & 49.24 & 11.47 \\ 
MSE  && 52.01 & 13.16\\ 
MSE  & \checkmark & 50.61 & 11.83  \\ 
\midrule
PiCIE  & & 54.08 & 14.11  \\
PiCIE  & \checkmark & 54.16 & 13.89  \\
PiCIE + H.  & & 54.65 & 14.32 \\
PiCIE + H.  &\checkmark & \textbf{54.75} & \textbf{14.77} \\
\bottomrule
\end{tabular}
}
\end{center}
\vspace{-2mm}
\caption{\textbf{Transfer learning results.} A new linear classifier has been trained on top of the learned embedding network. }
\label{tab:coco_linear}
\vspace{-6mm}
\end{table}

%% file: retraining_table.tex
\begin{table}
\begin{center}
\resizebox{\linewidth}{!}{
\begin{tabular}{lccccc}
Initialization   & Normalization &  Acc. & mIoU & C-Acc. & C-mIoU \\
\toprule
ImageNet  & & 75.48 & 44.69 & 55.82 &17.36 \\ 
ImageNet  & \checkmark & 74.74 & 43.44 & 57.24 & 31.51\\
\midrule
Modified DC  & & 75.25 & 44.37 & 55.16 & 18.43 \\ 
Modified DC  & \checkmark &  75.27 & 43.82 &  57.41 & 30.27 \\
IIC  & &75.16  &44.26  &56.07 & 20.32 \\
IIC  & \checkmark & 74.81 & 44.11  & 57.30 & 29.47 \\
\midrule
PiCIE  &  &  75.61 &44.40 &54.84 & 17.39    \\
PiCIE  & \checkmark & \textbf{76.02} &44.97 & \textbf{59.77} & \textbf{32.81 } \\
PiCIE + H.  & & 75.90 & \textbf{45.60} & 58.95 & 18.38 \\
PiCIE + H.  &\checkmark & 76.01 & 45.04 & 58.94 & 32.15\\
\bottomrule
\end{tabular}
}
\end{center}
\vspace{-2mm}
\caption{\textbf{Re-training results.} Trained networks are used as an initialization for standard supervised training. ``C-Acc.'' and ``C-mIoU'' are clustering results after supervised training. All models are trained from ImageNet-pretrained initialization. }
\vspace{-8mm}
\label{tab:coco_finetune}
\end{table}

%% file: iic_lambda.tex
\begin{table}
\begin{center}
\resizebox{0.8\linewidth}{!}{
\begin{tabular}{ccccccccc}
$\lambda$ & 1.0 & 1.25 & 1.5 & 1.75& 2.0& 3.0 \\
\toprule
Acc & 21.8 & 17.6 & 16.8 & 15.6 &16.4 & 16.6\\
mIoU & 6.7 & 7.0 & 6.4 & 6.0 &  6.5 & 6.5 \\
\bottomrule
\end{tabular}
}
\end{center}
\vspace{-2mm}
\caption{ IIC with different $\lambda$. }
\vspace{-10mm}
\label{tab:iic_lambda}
\end{table}

%% file: robust.tex
\begin{table*}
\begin{center}
\resizebox{\linewidth}{!}{
\begin{tabular}{ccccccccccc}
Brightness & Contrast & Saturation & Hue & Grayscale & Gaussian blur & Horizontal Flip & Random Crop & Accuracy & mIoU\\
\toprule
&&&&&&&& 48.09 & 13.84\\ 
\checkmark&&&&&&&& 47.98 & 13.59\\
&\checkmark&&&&&&& 48.08 & 13.63 \\ 
&&\checkmark&&&&&& 48.09 & 13.64 \\
&&&\checkmark&&&&& 48.09 & 13.65 \\
&&&&\checkmark&&&& 47.98 & 13.63 \\
&&&&&\checkmark&&& 48.03 & 13.59\\
\checkmark&\checkmark&\checkmark&\checkmark&\checkmark&\checkmark&&& 47.42 & 13.39 \\
&&&&&&\checkmark&& 47.61 & 13.71 \\
&&&&&&&\checkmark& 48.08 & 13.63\\ 
&&&&&&\checkmark&\checkmark & 47.60 & 13.76 \\ 
\checkmark&\checkmark&\checkmark&\checkmark&\checkmark&\checkmark&\checkmark&\checkmark& 46.28 & 13.16 \\ 
\bottomrule
\end{tabular}
}%
\end{center}
\vspace{-2mm}
\caption{We evaluate PiCIE with test-time augmentation where each transformation follows the same hyper-parameters as training, when applied. The result shows that PiCIE is robust to photometric and geometric transformations during inference. }
\vspace{-6mm}
\label{tab:robust}
\end{table*}